\documentclass[runningheads]{llncs}

\usepackage{etex}

\usepackage[utf8]{inputenc}
\usepackage{graphicx}
\usepackage{amsmath,amssymb} 
\usepackage{color}

\usepackage[noend]{algpseudocode}
\usepackage{algorithm}
\usepackage{algorithmicx}

\usepackage{lineno}
\let\doendproof\endproof
\renewcommand\endproof{~\hfill$\qed$\doendproof}

\usepackage{tikz}
\usetikzlibrary{decorations.pathreplacing,shapes,positioning,arrows}
\usepackage{textcomp}
\usepackage{booktabs}
\usepackage{array}
\newcolumntype{L}[1]{>{\raggedright\arraybackslash}p{#1}}
\usepackage{graphicx}
\usepackage{multirow}
\usepackage{caption,tabularx}
\usepackage{subcaption}
\usepackage{xspace}
\usepackage[square, sort, numbers]{natbib}
\usepackage{graphicx}
\usepackage{todonotes}
\usepackage{float}
\usepackage{pgfplots}
\usepackage{tabularx}
\usepackage{soul}
\usepackage{enumitem}

\usepackage{footmisc}

\usepackage{hyperref}
\usepackage{cleveref}

\newcommand{\func}[1]{{\tt #1}}
\newcommand{\elu}{\func{ELU}\xspace}
\newcommand{\selu}{\func{SeLU}\xspace}
\newcommand{\relu}{\func{ReLU}\xspace}
\newcommand{\swish}{\func{Swish}\xspace}
\newcommand{\sigmoid}{\func{Sigmoid}\xspace}
\newcommand{\tah}{\func{Tanh}\xspace}
\newcommand{\linear}{\func{Linear}\xspace}
\newcommand{\elish}{\func{ELiSH}\xspace}
\newcommand{\hardsigmoid}{\func{HardSigmoid}\xspace}
\newcommand{\sinus}{\func{Sin}\xspace}

\newcommand{\hardelish}{\func{HardELiSH}\xspace}
\newcommand{\softplus}{\func{Softplus}\xspace}

\DeclareMathOperator\caret{\raisebox{0.5ex}{$\scriptstyle\wedge$}}

\newcommand{\comment}[1]{}

\definecolor{BrightRoyalPurple}{rgb}{0.65, 0.05, 0.78}

\def\eg{e.g.,\xspace} 
\def\ie{i.e.,\xspace}

\newdimen\parboxheight

\makeatletter
\newcommand*{\org@iiiparbox}{}
\let\org@iiiparbox\@iiiparbox
\renewcommand*{\@iiiparbox}[2]{%
  \ifx\relax#2%
    \setlength{\parboxheight}{0pt}%
  \else
    \setlength{\parboxheight}{#2}%
  \fi
  \org@iiiparbox{#1}{#2}%
}
\makeatother

\begin{document}
\pagestyle{headings}
\mainmatter

\def\ACCV18SubNumber{333}  

\title{The Quest for the Golden Activation Function} 
\comment{peter: Indeed your title is the better one, however, for visibility all buzzwords should be in the title. Or, the title must be that presumptuous that everyone can recognize the idea just from the title ...}
\comment{Then maybe we should use my title and use your's as the subtitle so it would be both catchy and descriptive at the same time. Honestly, I like analogy with the Golden Fleece and how it would actually attract the readers and reviewers.}
\comment{Both would be too much! Finally, it's your paper}

\titlerunning{~}
\authorrunning{~}

\author{Mina Basirat and Peter M. Roth}
\institute{Graz University of Technology \\
Institute of Computer Graphics and Vision \\
\{mina.basirat, pmroth\}@icg.tugraz.at}

\maketitle
\begin{abstract}

Deep Neural Networks have been shown to be beneficial for a variety of tasks, in 
particular allowing for end-to-end learning and reducing the requirement for 
manual design decisions. However, still many parameters have to be chosen in 
advance, also raising the need to optimize them. One important, but often 
ignored system parameter is the selection of a proper activation function. Thus, 
in this paper we target to demonstrate the importance of activation functions in 
general and show that for different tasks different activation functions might 
be meaningful. To avoid the manual design or selection of activation functions, 
we build on the idea of genetic algorithms to learn the best activation function 
for a given task. In addition, we introduce two new activation functions, \elish 
and \hardelish, which can easily be incorporated in our framework. In this way, 
we demonstrate for three different image classification benchmarks that 
different activation functions are learned, also showing improved results 
compared to typically used baselines. 

\end{abstract}

\section{Introduction}

Deep Neural Networks (DNNs) (see \eg \cite{pmroth:goodfellow16,pmroth:lecun15}) 
have recently become very popular and are now successfully applied for a wide 
range of applications. However, as more complex and  deeper networks are of 
interest, strategies are required to make neural network training more efficient 
and more stable. While for instance initialization (\eg 
\cite{Sutskever:2013:IIM:3042817.3043064,Mishkin2015AllYN}) and normalization 
techniques (\eg \cite{laurent16}) are well studied, an also relevant and 
important factor is often neglected: the role of activation functions (AFs). 
Even though recent work demonstrated that AFs are of high relevance (see \eg 
\cite{DBLP:conf/nips/KlambauerUMH17, Ramachandran18, 
DBLP:journals/corr/ClevertUH15, Silu, softplus,hardsigmoid}), due to is 
simplicity and reliability most deep learning approaches use Rectified Linear 
Units (\relu) \cite{Nair:2010:RLU:3104322.3104425} as nonlinear activation 
functions. 

Initially, due to their universal approximation properties the research in this 
field was mostly concentrated on squashing functions such as \sigmoid and \tah  
\cite{Hornik:1991:ACM:109691.109700}. However, training DNNs using such 
functions suffers from the vanishing gradient problem \cite{vanishing}.
%
%
To overcome this problem, various non-squashing functions were introduced, where 
the most notable example is Rectified Linear Unit (\relu) 
\cite{Nair:2010:RLU:3104322.3104425}. In particular, as the derivative of 
positive inputs in \relu are one, the gradient cannot vanish. In contrast, as 
all negative values are mapped to zero, there is not information flow in DNNs 
for negative values. This problem is known as dying \relu. 

To deal with this problem, various generalizations of \relu such \func{Leaky} 
\relu \cite{Maas13rectifiernonlinearities} have been proposed. 
%
%
Similarly, Exponential Linear Units (\elu) \cite{DBLP:journals/corr/ClevertUH15} 
do not only eliminate the bias shift in the succeeding layers, but also push the 
mean activation value towards zero by returning a bounded exponential value for 
negative inputs. Although, showing competitive results, \elu is not backed by a 
very strong theory. A theoretically proven extension, Scaled Exponentiation 
Linear Unites (\selu) \cite{DBLP:conf/nips/KlambauerUMH17}, makes DNN learning 
more robust. It fact, it is shown that the proposed self-normalizing network 
converges towards a normal distribution with zero mean and unit variance. 

A different direction was pursued in \cite{Ramachandran18}, finally introducing 
the \swish activation function.  Different search spaces are created by varying 
the number of core units used to construct the activation function, and an RNN 
is trained to search the state space for a novel activation function. The 
proposed approach shows competitive results for both shallow and deep neural 
networks. Recently, a theoretic proof and justification for the design have been 
given in \cite{hayou18}, showing that \swish propagates information better than 
\relu.

In this way, existing approaches to estimate activation functions for DNN 
learning are lacking theoretical foundation, are based on complex theory, which 
is hard to understand in the context of practical applications, or are based on 
inefficient search schemes, which still require to manually set several 
parameters. To overcome these problems, as first contribution, we propose an 
approach based on ideas of Genetic Algorithms \cite{GA-intro}. 

In particular, building on neuro-evolutionary algorithms \cite{Neuro-intro}, 
starting from simple initial activation functions more and more complex 
functions can be obtained over time, which are better suited for a given task. 
In contrast, to brute-force search strategies, the search space is explored in a 
more efficient way, drastically reducing the training effort. In addition, we 
propose to define piece-wise functions, better representing the desired 
properties. In fact, this idea can easily be included in the proposed learning 
framework.  

As second contribution, based on recent theoretical findings \cite{hayou18}, we 
introduce two new activation functions, namely, \elish and \hardelish, which 
have shown to be competitive compared to existing approaches as well as very 
useful within the proposed framework. To demonstrate the benefits of our learned  
activation functions, we applied our approach for three different object 
classification benchmark data sets, varying in size and complexity, and run it 
using two different network architectures. The results clearly demonstrate that 
using the proposed approach better results can be obtained as well as that for 
different tasks different activation functions are useful.

The reminder of the paper is structured as follows: First, in 
Sec.~\ref{sec:related}, we discuss the related work in the context of Genetic 
Algorithms for Neural Networks. Next, in 
Secs.~\ref{sec:elish}~and~\ref{sec:lcaf} we introduce our new activation 
functions and the new neuro-evolutionary algorithm for learning task-specific 
activation functions. Then, in Sec.~\ref{sec:results}, we give a detailed 
experimental evaluation of our approach and discuss the findings. Finally, in 
Sec.~\ref{sec:conclusion} we summarize and conclude our work. 
 
\pagebreak

\section{Related Work}
\label{sec:related}

Neuroevolution, \ie applying evolutionary algorithms (EAs) in the optimization  
of DNNs \cite{Whitley01anoverview}, is a vital field of research. In general, 
there are two main directions. First, optimizing training parameters such as 
hyper-parameters \citep{loshchilov2016cma} or weights 
\cite{Montana:1989:TFN:1623755.1623876,184facbcadfa44edab389b49b5130ca5}. In the 
latter case, in contrast to methods like gradient decent, also global optima can 
be estimated. Second, evolving an optimal DNN topology, which, however, is not 
straightforward. Therefore, existing approaches follow two strategies: 
constructive \cite{constructivegrow} and destructive 
\cite{Hancock92pruningneural}. Constructive methods start from a simple topology 
and gradually increase the complexity until an optimality criterion is 
satisfied. In contrast, destructive approaches start from an initially complex 
topology and incrementally reduce the unnecessary structures.

Recently, co-evolution of topology and weights (TWEANNs) has shown to be more 
effective and efficient. The most successful related approach NEAT 
\cite{Stanley:2002:ENN:638553.638554}. NEAT follows the constructive strategy 
and gradually evolves a simple DNN topology towards unbounded complexity by 
adding nodes and connections between them while preserving the optimality of 
topology. Due to its success, there have been several extensions of NEAT.  For 
instance, in \cite{DBLP:journals/corr/MiikkulainenLMR17} two extensions, 
DeepNEAT and CoDeepNEAT, have been proposed. In contrast to NEAT, in DeepNEAT a 
node represents a layer and consists of a table of hyper-parameters (\ie number 
of neurons ) related to it. In CoDeepNEAT, two populations (modules and 
blueprints) are initialized separately, where a module is a graph and represents 
a shallow DNN. A blueprint has also a graph structure and consists of nodes 
pointing out to specific module species. Both modules and blueprints evolve in 
parallel, and, finally, the modules and blueprints are combined to build up the 
topology of the DNN.

Similarly, \cite{suganuma2017genetic} explored a CNN architecture via Cartesian 
Genetic programming (CGP) for image classification, where also high level 
functions such as convolution or pooling operations are implemented. Recently, 
\cite{suganuma2018genetic} proposed a constructive hierarchical genetic 
representation approach for evolving DNN topologies. Initialized with small 
populations of primitives such as convolutional and pooling operations at the 
bottom of the hierarchy, the topology gets more and more complex by adding 
evolved primitives into graph structure.

So far most attentions have been drawn to TWEANNs, however, we are interested in 
evolving activation functions, which was only of limited interest up to now. The 
ideas closest to ours are Hyper-NEAT \cite{stanley2009hypercube} and HA-NEAT 
\cite{hagg2017evolving}. Hyper-NEAT is a NEAT extension to evolve connective 
compositional pattern-producing networks (CPPNs) to estimate the weights of the 
ANN. In fact, the geometry of patterns can be represented by a composition of 
functions. Similarly, HA-NEAT \cite{hagg2017evolving} extends NEAT to evolve 
activation functions of neurons, topology, and weights, resulting in a 
heterogeneous network. In contrast, we fixed the topology and evolved the 
piece-wise activation functions on layer level. The proposed candidate solutions 
are more complicated (advanced) than those of HA-NEAT. More importantly, the 
complexity of evolved activation functions is, in contrast to HA-NEAT, 
unbounded. Nevertheless, it is possible to evolve our idea along with topology.

\pagebreak

\section{ELiSH: Exponential Linear Sigmoid SquasHing}
\label{sec:elish}

Based on recent findings in \cite{Ramachandran18} and \cite{hayou18}, in the 
following, we introduce two new activation functions, \elish and \hardelish, 
which will then also be applied in our evolutional framework. In particular, the 
design of these activation functions was motivated by the recently proposed 
\swish activation function  \cite{Ramachandran18}:

\begin{equation}
\label{eq:swish}
y(x) =x/(1+e^{-x}).
\end{equation}
 
In fact, \swish  possess various properties desirable for activation functions. 
In fact, the function is unbounded above, bounded below, non-monotonic, and 
smooth \cite{Ramachandran18}. In addition, in \cite{hayou18} it was also shown 
that it provides a good information flow through a DNN.  In general, 
\cite{Ramachandran18} and \cite{hayou18} identified a family of activation 
functions in the form of $y(x) = x \cdot \textrm{sigmoid}(x)$, which improves 
the information propagation and does not suffer form the vanishing gradient 
problem. In this way, we introduce the new Exponential Linear Sigmoid SquasHing 
(\elish) activation function:

\begin{equation}
  \label{eq:elish}
  y(x)=
  \begin{cases}
     x/(1+e^{-x}) & x\geq 0\\
    (e^{x}-1)/(1+e^{-x}) & x< 0.\\
  \end{cases}
\end{equation}

From Eq.~(\ref{eq:elish}) it is clear that \elish shares the properties of 
\swish, as its negative part is a multiplication of \elu and \sigmoid, while 
sharing the same positive part with \swish. Similarly, we introduce \hardelish 
as a multiplication of \hardsigmoid and \elu in negative part and \hardsigmoid 
and \linear in positive part\footnote{See \Cref{table:1} for the definition of 
these functions.}:

\begin{equation}
  \label{eq:harelish}
  y(x)=
  \begin{cases}
     x\times\max(0,\min(1,(x+1)/2)) & x\geq 0\\
     (e^{x}-1)\times\max(0,\min(1,(x+1)/2)) & x< 0.\\
  \end{cases}
\end{equation}

\begin{figure}
\centering
\resizebox{0.75\textwidth}{!}
{
\begin{tabular}{cc}
    \begin{tikzpicture}[scale=2/3]
\begin{axis}
    [xmin=-5, xmax=4,
    ymin=-2, ymax=6,
    x=0.75cm,
    y=0.75cm,
    axis lines=center,
    axis on top=true]

    \addplot [mark=none,draw=red,ultra thick,domain=0:4] { max(0, min(1, (x+1)/2))*x};
    \addplot [mark=none,draw=red,ultra thick,domain=-10:0] {( exp(x)-1) * max(0, min(1, (x+1)/2))};
    \addplot [mark=none,draw=blue,dotted,ultra thick,domain=-2:0] {2.5*(exp(x)-1)+exp(x)*max(0, min(1, (x+1)/2))};
     \addplot [mark=none,draw=blue,dotted,ultra thick,domain=-6:-2] {exp(x)*max(0, min(1, (x+1)/2))};
    \addplot [mark=none,draw=blue,dotted,ultra thick,domain=0:2] { max(0, min(1, (x+1)/2))+2.5*x};
    \addplot [mark=none,draw=blue,dotted,ultra thick,domain=2:4] { max(0, min(1, (x+1)/2))};
\end{axis}
\end{tikzpicture}  &
    \begin{tikzpicture}[scale=2/3]
\begin{axis}
    [xmin=-5, xmax=5,
    ymin=-2, ymax=6,
    x=0.75cm,
    y=0.75cm,
    axis lines=center,
    axis on top=true]
    \addplot [mark=none,draw=red,ultra thick,domain=0:5] {x*1/(1+exp(-x))};
     \addplot [mark=none,draw=red,ultra thick,domain=-10:0]{(exp(x)-1)*1/(1+exp(-x))};
    \addplot [blue, dotted, ultra thick, domain=0:5]{(exp(-x)/(1+exp(-x))^2)*x+1/(1+exp(-x))};
     \addplot [blue, dotted, ultra thick, domain=-10:0]{(exp(-x)/(1+exp(-x))^2)*x+1/(1+exp(-x))};
\end{axis}
\end{tikzpicture}
     \\
    $y(x) =\hardelish$ &
    $y(x) = \elish$ \\
\end{tabular}
}

\caption{The proposed new activation functions (red), \hardelish and \elish, and 
their derivatives (blue dotted), respectively.} \label{plot:elish}
\end{figure}
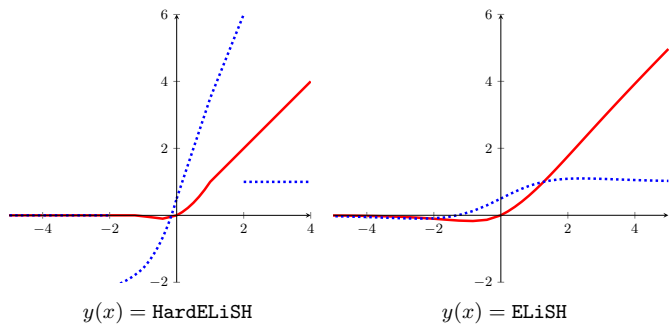

Moreover, we would like to take advantage of compositional functions. For 
example in \swish, \sigmoid improves the information flow and  \linear avoids a 
vanishing gradient, which is also the main motivation to design \elish and 
\hardelish. Both activation functions and their derivatives are shown in 
\Cref{plot:elish}.

\section{Evolving Piece-wise Activation Functions}
\label{sec:lcaf}

The goal of this work is estimate non-linear activation functions better suited 
for specific tasks. To this end, we build on two ideas. First, as negative and 
positive inputs have a different influence on learning, we propose to use split 
activation functions. Second, as the search space can be very large, we propose 
to build on the ideas of genetic algorithms to allow for a more efficient 
search.

\subsection{Genetic Algorithms}
\label{sec:ga}

Genetic Algorithms (GA) (see \eg \cite{DeJong2006a}) can be seen a 
population-based meta-heuristic to solve problems in the field of stochastic 
optimization. In particular, we are given a large set of candidate solutions, 
referred to as \emph{population}, but we do not now how to approach the global 
optimum. In this way, the main idea is to evolve a population to a better 
solution.  

The evolution typically starts from a population consisting of randomly selected 
candidate solutions. These are called \emph{individuals} and are described by a 
set of properties (\emph{gens}), which can be altered by three bio-inspired 
operations: (a) \emph{selection}, (b) \emph{crossover}, and (c) \emph{mutation}. 
Selection is the simple process of selecting individuals according to their 
fitness. In contrast, crossover is a stochastic operator, exchanging information 
between two individuals (often called \emph{parents}: \emph{mom} and \emph{dad}) 
to form a new offspring. Similarly, mutation is also a stochastic operator that 
helps to increase the diversity of the population by randomly choosing one or 
more genes in an offspring and changing them.  

Then, an iterative process, where an iteration is referred to as 
\emph{generation}, the fitness of each individual is evaluated. Based on their 
fitness, we select a set of parents solutions for breeding. Subsequently, we 
apply breeding operators on pairs of individuals to generate new pairs of 
offsprings. Eventually, we update the population with the set of parents and 
bred offsprings. This process is repeated until a predefined number of 
generations or an optimality criterion is met.

\subsection{Genetic Operators for Activation Functions}

\begin{figure}[!h]
\centering
\begin{tikzpicture}[scale=4/5, transform shape]
    \draw (0,0) rectangle node{activation left} (2.5,0.75);
    \draw (2.5,0) rectangle node{activation right} (5,0.75);
    \draw[decoration={brace},decorate](0,1)--node[blue,above=0.1]{Activation function} (5,1);
    \draw[decoration={brace,mirror},decorate](0,-0.25) -- node[blue,below=0.1] {Gene} (2.5,-0.25);
\end{tikzpicture}
\caption{An individual in the population of our GA.}
\label{fig:individual}
\end{figure}
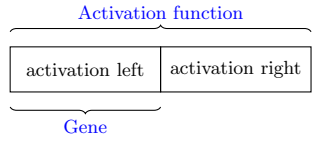

In our case, targeting to evolve piecewise activation functions, our populations 
consists of individuals representing an activation function, where a gene is 
either the left or the right part of an activation function. This is illustrated 
in \Cref{fig:individual}.

To evolve activation functions as described above, we need first to introduce 
new operators, representing our problem, \emph{Inheritance} and \emph{Hybrid}. 
The latter operator combines the parents' activation functions by means of 
mathematical operators. When applying the crossover operator we stochastically 
choose between Inheritance and Hybrid as shown in \Cref{alg:method:crossover}. 

\vspace{-0.5cm}
\subsubsection{Inheritance}

\begin{figure}
\centering
\begin{subfigure}{0.5\textwidth}
\begin{tikzpicture}[scale=4/5, transform shape]
    \draw (0,0) rectangle node{mom's left} (2,0.75);
    \draw (2,0) rectangle node{dad's right} (4,0.75);
\end{tikzpicture}
\parbox[c]{.4\linewidth}{\caption{Offspring 1.}}
\label{fig:crossover:1:child1}
\end{subfigure}
\begin{subfigure}{0.49\textwidth}
\centering
\begin{tikzpicture}[scale=4/5, transform shape]
    \draw (0,0) rectangle node{dad's left} (2,0.75);
    \draw (2,0) rectangle node{mom's right} (4,0.75);
\end{tikzpicture}
\parbox[c]{.4\linewidth}{\caption{Offspring 2.}}
\label{fig:crossover:1:child2}
\end{subfigure}
\caption{Inheritance Crossover.}\label{fig:crossover:1}
\end{figure}
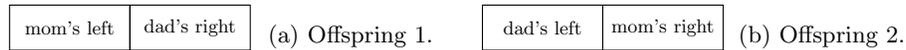

The Inheritance operator is intended to inherit genes from both parents.  The 
first (second) offspring inherits its left activation function from the mom 
(dad), and its right activation function from the dad (mom). Thus, the operator 
is defined in a similar way as a one point crossover operator, however, the 
cuttoff point is predetermined (\ie we are dealing with functions). This is 
illustrated in \Cref{fig:crossover:1}.

\vspace{-1.0cm}

\subsubsection{Hybrid Crossover}

\begin{figure}
\centering
\begin{subfigure}{\textwidth}
\centering
\begin{tikzpicture}[scale=4/5, transform shape]
    \draw (0,0) rectangle node{$op_1$, mom's left, dad's left} (4.5,0.75);
    \draw (4.5,0) rectangle node{$op_2$, mom's right, dad's right} (9,0.75);
\end{tikzpicture}
\parbox[b]{.3\linewidth}{\caption{Offspring 1.}}
\label{fig:crossover:2:child1}
\end{subfigure}

\begin{subfigure}{\textwidth}
\centering
\begin{tikzpicture}[scale=4/5, transform shape]
    \draw (0,0) rectangle node{$op_1$, dad's left, mom's left} (4.5,0.75);
    \draw (4.5,0) rectangle node{$op_2$, dad's right, mom's right} (9,0.75);
\end{tikzpicture}
\parbox[c]{.3\linewidth}{\caption{Offspring 2.}}
\label{fig:crossover:2:child2}
\end{subfigure}
\caption{Hybrid crossover operator, where $op_1$ and $op_2$ are chosen randomly.
}\label{fig:crossover:2}
\end{figure}
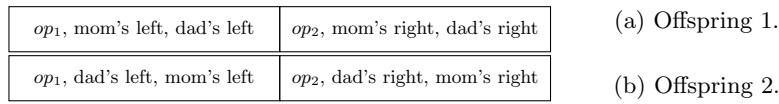

The Hybrid crossover operator is proposed to combine multiple activation 
functions. As for Inheritance crossover, the cutoff point is fixed. Using a 
randomly selected mathematical operator, the first (second) offspring combines 
mom's and dad's (dad's and mom's) negative part of the activation function to 
form its own negative part. Subsequently, the first (second) offspring's 
positive part of the activation function is formed via a combination of mom's 
and dad's (dad's and mom's) positive part. This is illustrated in 
\Cref{fig:crossover:2}.

\vspace{-0.5cm}
\subsubsection{Mutation Operator}
\label{sec:operators}

The mutation operator randomly chooses a gene and then replaces it with a 
randomly selected predefined activation function. In fact, this operator helps 
our GA algorithm to keep exploring the search space for new activation 
functions. This is illustrated in Figure~5. 


\begin{figure}
\label{fig:mutation}
\centering
\begin{tikzpicture}[scale=4/5, transform shape]
    \draw (0,0) rectangle node (A){mom's left} (2,0.75);
    \node [left=1cm of A] (B) {\begin{tabular}{c}Selected Activation:\\\textcolor{blue}{\relu}\end{tabular}};
    \draw (2,0) rectangle node{dad's right} (4,0.75);
    \draw [decoration={brace},decorate](0,0.85) -- node[blue,above=0.1] {Selected Gene} (2,0.85);
    \draw [->] (B) -- (A);
\end{tikzpicture}
\parbox[][10pt][t]{.4\linewidth}
{\vspace{-10pt}
\caption{Mutation operator.}}

\end{figure}
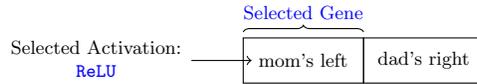

\vspace{-1cm}

\subsection{Evaluating an Activation Function}

The Hybrid crossover operator results in a hybrid activation function that we 
evaluate by parsing according to the following grammar: 
\begin{align}
    expression :=& f~|~operation,~expression,~expression \notag  \\
    operation :=& +~| - | \times |~/~|\caret| \min | \max | ~f\circ g \notag \\
    f :=&~\elish~|~\hardelish~|~\swish~|~\relu~|~\elu~|~\selu~| \dots,
    \label{eq:function_list}
\end{align}
where $f$ represents the set of candidate solutions. The list is not fixed, and 
we can easily add additional $operations$ and candidate solutions $f$.

\vspace{0.35cm}
\noindent \emph{Example:} Given an activation function generated by Hybrid crossover:
\[
    (\max:(+:(\min:\elu:\relu):\swish):(\times:\elu:\linear))\;.
\]

\noindent Using Eq.~\eqref{eq:function_list}, we parse above activation function 
as shown in \Cref{fig:parsetree} to compute the equivalent infix expression: 

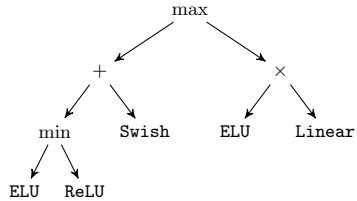
\begin{figure}
    \centering
    \begin{tikzpicture}[scale=0.8,transform shape,
    ->,>=stealth',level/.style={sibling distance = 3cm/#1,level distance = 1cm}] 
    \node  {max}
        child{ node {$+$}
                child{ node {min} 
                        child{node {\elu}}
                        child{node {\relu}}}
                child{ node {\swish } } }
        child{ node {$\times$}
                child{ node {\elu}}
                child{ node {\linear}}};
    \end{tikzpicture}
\caption{The parse tree of $\max((\min(\elu,\relu)+\swish), (\elu \times \linear))$.
\label{fig:parsetree}}
\end{figure}

\vspace{-0.75cm}

\subsection{Learning Activation Functions}
\label{sec:laf}

Having defined the newly defined genetic operators and having explained the 
evaluation, we can now introduce the overall evolutionary approach, which is 
summarized in \Cref{alg:ga}. Initially, we generate a population of random 
activation functions (\Cref{alg:ga:pop:init}). Next, using the evaluate operator 
(\Cref{alg:ga:fitness:init}), the fitness of each individual is determined by 
train and test performance of a DNN. Indeed, the DNN uses an individual as its 
activation function. Then, we select a set of parent activation functions based 
on their fitness for breeding (\Cref{alg:ga:select}). To generate new activation 
functions (\Cref{alg:ga:crossover}), we apply a new crossover operator as 
defined in \Cref{sec:operators} and afterwards the mutation operator. Similarly, 
we update our population with the set of parents and bred offsprings and 
continue to the next generation. This procedure is iterated until a 
pre-pre-defined optimality criterion is met.

\begin{figure}
\centering

\resizebox{\linewidth}{!}{%

\begin{tabular}{cm{5mm}c}

\begin{minipage}{0.6\linewidth}
\begin{algorithm}[H]
    \caption{Genetic Algorithm} \label{alg:ga}
    \begin{algorithmic}[1]
     \Procedure{GA}{population-size}
     \State{population $\gets \emptyset$}
     \State{population $\gets$ \textsc{Initialize}(population-size)}
     \label{alg:ga:pop:init}
     \State{\textsc{Evaluate}(population)}
     \label{alg:ga:fitness:init}
     \Repeat
     \label{alg:ga:generation:start}
        \State{children $\gets \emptyset$}
        \State{parents $\gets$ \textsc{Select}(population, $15\%$)}\label{alg:ga:select}
        \For{$i \leq$ (population-size $-|$parents$|)/2$}\label{alg:ga:breeding:start}
            \State{increment $i$ by one}
            \State{$\langle$mom, dad$\rangle \in$ parents $\times$ parents}
            \State{offsprings $\gets$ \textsc{Crossover}(mom, dad)}\label{alg:ga:crossover}
            \For{offspring $\in$ offsprings}
                \State{offspring $\gets$ \textsc{Mutate}(offspring)}
            \EndFor
            \State{\textsc{Evaluate}(offsprings)}
            \label{alg:ga:fitness:offsprings}
            \State{children $\gets$ children $\cup$ offsprings}
        \EndFor\label{alg:ga:breeding:end}
        \State{population $\gets$ parents $\cup$ children}
        \label{alg:ga:pop:update}
    \Until{termination condition}
    \label{alg:ga:generation:end}
    \State{\Return population}
    \EndProcedure
    \end{algorithmic}
\end{algorithm}
\end{minipage}

&
&

\begin{minipage}{0.65\linewidth}

\begin{algorithm}[H]
    \caption{Selection Operator} \label{alg:ga:selection}
    \begin{algorithmic}[1]
    \Procedure{Select}{population, $n$}
    \State{parents $\gets$ top $n\%$ of population}
    \For{individual $\in$ population $-$ parents}
       \State{$c \gets$ toss a coin}
       \If{$c$ is heads}
       \State{parents $\gets$ parents $\cup$ \textsc{Mutate}(individual)}
       \EndIf
    \EndFor
    \EndProcedure
    \end{algorithmic}
\end{algorithm}

\begin{algorithm}[H]
    \caption{} \label{alg:method:crossover}
    \begin{algorithmic}[1]
    \Procedure{Crossover}{mom, dad}
    \State{$c \gets$ toss a coin}
    \If{$c$ is heads}
        \State{\Return \textsc{Inheritance}(mom, dad)}
    \EndIf
    \State{\Return \textsc{Hybrid}(mom, dad)}
    \EndProcedure
    \end{algorithmic}
\end{algorithm}

\end{minipage}

\end{tabular}

}
\end{figure}

\section{Experimental Results}
\label{sec:results}

The purpose of our experiments is threefold. First, we would like to show that 
for different tasks different choices of activation functions are meaningful. 
Second, we demonstrate the generality of the evolved activation functions by 
applying them using a different architecture. Third, we show that the best 
performing activation functions are similar, representing a specific 
characteristics of the data. In particular, we run experiments on three 
different object classifications benchmarks (\ie  
CIFAR-10\footnote{\label{note1}\url{https://www.cs.toronto.edu/~kriz/cifar.html}}, 
CIFAR-100\footref{note1}, and Tiny 
ImageNet\footnote{\url{https://tiny-imagenet.herokuapp.com/}}), differing in 
number of classes, number of samples, and complexity, and by using two different 
DNN architectures (\ie  preactivation-ResNet
\cite{DBLP:conf/eccv/HeZRS16} and VGG \cite{Simonyan15}).


\subsection{Experimental Setup and Implementation Details}
\label{sec:exp-setup}

Similar to \cite{Ramachandran18}, we run our Genetic Algorithm based learning 
strategy on more shallow architectures, \ie  ResNet38 for CIFAR-10 and  ResNet20 
for CIFAR-100 and Tiny ImageNet. The thus explored activation functions are then 
used for training deeper networks, \ie  Resnet56 \cite{DBLP:conf/eccv/HeZRS16}. 
In addition, to demonstrate that the obtained activation functions are of more 
general interest, the selected functions are additionally applied for training 
classifiers based on VGG-16 \cite{Simonyan15}. To this end, we used the default 
parameters for both architectures. However, to avoid random effects, all 
networks have been initialized using the same initialization 
\cite{He:2015:DDR:2919332.2919814}; moreover, to keep the computational effort 
feasible\footnote{The experiments were carried out on a standard PC (Core-i7, 
64GB RAM) with two Titan-X GPUs attached.}, the batch size was set to $32$.

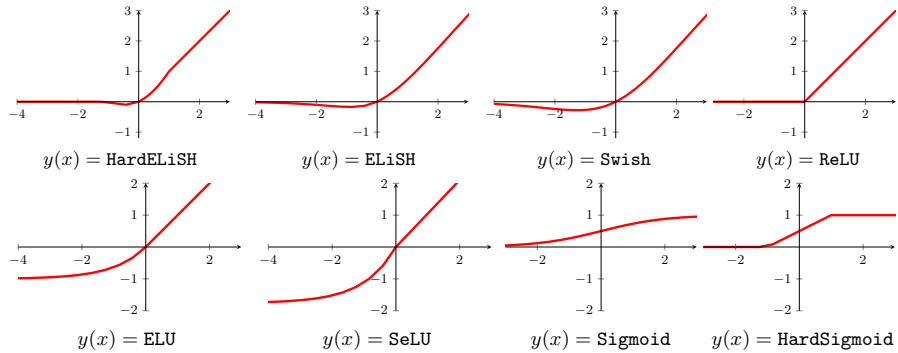
\begin{figure}
\centering
\resizebox{\textwidth}{!}
{
\begin{tabular}{cccc}
    \begin{tikzpicture}[scale=2/3]
\begin{axis}
    [xmin=-4, xmax=3,
    ymin=-1.2, ymax=3,
    x=0.75cm,
    y=0.75cm,
    axis lines=center,
    axis on top=true]

    \addplot [mark=none,draw=red,ultra thick,domain=0:4] { max(0, min(1, (x+1)/2))*x};
    \addplot [mark=none,draw=red,ultra thick,domain=-10:0] {( exp(x)-1) * max(0, min(1, (x+1)/2))};
\end{axis}
\end{tikzpicture} &
    \begin{tikzpicture}[scale=2/3]
\begin{axis}
    [xmin=-4, xmax=3,
    ymin=-1.2, ymax=3,
    x=0.75cm,
    y=0.75cm,
    axis lines=center,
    axis on top=true]
    \addplot [mark=none,draw=red,ultra thick,domain=0:4] {x*1/(1+exp(-x))};
    \addplot [mark=none,draw=red,ultra thick,domain=-10:0] {(exp(x)-1)*1/(1+exp(-x))};
\end{axis}
\end{tikzpicture} &
    \begin{tikzpicture}[scale=2/3]
\begin{axis}
    [xmin=-4, xmax=3,
    ymin=-1.2, ymax=3,
    x=0.75cm,
    y=0.75cm,
    axis lines=center,
    axis on top=true]

    \addplot [mark=none,draw=red,ultra thick,domain=0:4] {x*1/(1+exp(-x))};
    \addplot [mark=none,draw=red,ultra thick,domain=-4:0] {x*1/(1+exp(-x))};
\end{axis}
\end{tikzpicture}&
    \begin{tikzpicture}[scale=2/3]
    \begin{axis}
        [xmin=-3, xmax=3,
        ymin=-1.2, ymax=3,
        x=0.75cm,
        y=0.75cm,
        axis lines=center,
        axis on top=true]
        \addplot [mark=none,draw=red,ultra thick,domain=0:4] {\x};
        \addplot [mark=none,draw=red,ultra thick,domain=-10:0] {0};
\end{axis}
\end{tikzpicture} 
    \\
    $y(x) = \hardelish$
    &
    $y(x) =\elish$ 
    &
    $y(x) =\swish$ 
    &
    $y(x) = \relu$
\end{tabular}
}
\resizebox{\textwidth}{!}{
\begin{tabular}{cccc}
    \begin{tikzpicture}[scale=2/3]
\begin{axis}
    [xmin=-4, xmax=3,
    ymin=-2, ymax=2,
    x=0.75cm,
    y=0.75cm,
    axis lines=center,
    axis on top=true]

    \addplot [mark=none,draw=red,ultra thick,domain=0:4] {\x};
    \addplot [mark=none,draw=red,ultra thick,domain=-10:0] {exp(\x)-1};
    
    
\end{axis}
\end{tikzpicture} &
    \begin{tikzpicture}[scale=2/3]
\begin{axis}
    [xmin=-4, xmax=3,
    ymin=-2, ymax=2,
    x=0.75cm,
    y=0.75cm,
    axis lines=center,
    axis on top=true]
    \addplot [mark=none,draw=red,ultra thick,domain=0:4] {1.0507009873554804934193349852946*x};
    \addplot [mark=none,draw=red,ultra thick,domain=-10:0] {1.0507009873554804934193349852946*1.6732632423543772848170429916717*(exp(\x)-1)};
\end{axis}
\end{tikzpicture} &
    \begin{tikzpicture}[scale=2/3]
\begin{axis}
  [xmin=-3, xmax=3,
    ymin=-2, ymax=2,
    x=0.75cm,
    y=0.75cm,
    axis lines=center,
    axis on top=true]
    \addplot [mark=none,draw=red,ultra thick,domain=0:6] {1/(1+exp(-x))};
    \addplot [mark=none,draw=red,ultra thick,domain=-10:0] {1/(1+exp(-x))};
\end{axis}
\end{tikzpicture}&
    \begin{tikzpicture}[scale=2/3]
\begin{axis}
    [xmin=-3, xmax=3,
    ymin=-2, ymax=2,
    x=0.75cm,
    y=0.75cm,
    axis lines=center,
    axis on top=true]
    \addplot [mark=none,draw=red,ultra thick,domain=0:6]{(  max(0, min(1, (x + 1)/2))};
    \addplot [mark=none,draw=red,ultra thick,domain=-10:0] {( max(0, min(1, (x + 1)/2))};
\end{axis}
\end{tikzpicture} 
    \\ 
    $y(x)=\elu$ 
    &
    $y(x)=\selu$
    &    
    $y(x)=\sigmoid$
    &
    $y(x)=\hardsigmoid$
\end{tabular}
}
\caption{Piecewise activation functions as defined in \Cref{table:1}.}
\label{plot:piece-wise activation functions}
\end{figure}

Our implementation for evolutionary learning builds on 
DeepEvolve\footnote{\url{https://github.com/jliphard/DeepEvolve}},  a 
neuroevolution framework developed to explore the optimal DNN topology for a 
given task. In our case, we fixed the DNN topology and defined the search space 
based on the activation functions. Throughout all experiments, we used a 
population size of $40$ and evolved the population over $8$ generations. The 
considered candidates for the initial population are shown in \Cref{table:1} and 
\Cref{plot:piece-wise activation functions}.

\vspace{-0.5cm}
\begin{table}[!h]
\caption{Candidate piece-wise activation functions. Please note, by $y(x<0)$ and $y(x\geq 0)$ we indicate the (left and right)-piece, respectively.}
\label{table:1}
\centering
\begin{tabular}{rm{1.8cm}l}
\toprule
\multicolumn{2}{c}{Activation Function}
&
\multicolumn{1}{c}{Expression}
\\
\midrule
1.
&
\hardelish
&
$
y(x<0) = \max(0,\min(1,(x+1)/2))\times (e^{x}-1)
$
\\ 
&
&
$y(x\geq0)=x\times \max(0,\min(1,(x+1)/2))$
\\

2.
&
\elish 
&
$
y(x<0) =(e^{x}-1)/(1+e^{-x})\text{ and } y(x\geq0)=x/(1+e^{-x})
$
\\
3.
&
\swish 
&
$
y(x) =x/(1+e^{-x})
$
\\
4.
&
\relu 
&
$
y(x)=\max(x,0)
$
\\
5.
&
\elu
&
$
y(x<0) = e^x-1 \text{ and } y(x\geq0)=x
$
\\
6.
&
\selu
&
$
y(x<0) = \lambda\alpha(e^x-1) \text{ and } y(x\geq0)=\lambda x
$
\\
7.
&
\softplus
&
$
y(x) = \ln(1+e^x)
$
\\
8.
&
\hardsigmoid
&
$
y(x) = \max(0,\min(1,(x+1)/2))
$
\\
9.
&
\sigmoid
&
$
y(x) = 1/(1+e^{-x})
$
\\
10.
&
\sinus
&
$
y(x) = \sin(x)
$
\\
11.
&
\linear
&
$
y(x) = x
$
\\
\bottomrule
\end{tabular}
\end{table}

\vfill

\clearpage

\subsection{Quantitative Results}
\label{sec:qres}


First, we evolved a set of candidate activation functions using our GA-based 
approach using ResNet-38 and used the evolved activation functions using 
ResNet-56 on CIFAR-10. The thus obtained results in terms of classification 
accuracy for the best performing solutions are shown in \Cref{table:2}. In 
addition, we give a comparison to three different baselines, namely \relu, \elu, 
and \selu, which have proven to work well for a wide range of applications.  
It can be seen from \Cref{table:2} that the best results can be obtained using 
\hardelish ($93.13\%$) and the activation function consisting of a combination 
of a multiplication of \hardelish and \swish in the positive part and \swish in 
the negative part ($93.02\%$). In general, it can be recognized that the top 6 
evolved activation functions are outperforming the baselines.

\vspace{-0.5cm}

\begin{table}
\centering
\caption{Performance of top six explored functions for Cifar10-ResNet56.}
\label{table:2}
\begin{tabular}{lcl}
\toprule
&
\multicolumn{1}{c}{Accuracy}
&
\multicolumn{1}{c}{Activation Function}
\\
\midrule

1.
&
93.13\%
&
$
y(x<0) = \hardelish
$
\\
2.
&
93.02\%
&
$
y(x<0) = \hardelish \circ \swish \text{ and } y(x\geq0)~=\swish 
$
\\
3.
&
92.89\%
&
$
y(x<0) = \sinus\text{ and } y(x\geq0)=\swish+\swish
$
\\
4.
&
92.83\%
&
$
y(x) =\swish
$
\\
5.
&
92.83\%
&
$
y(x) =\elish
$
\\
6.
&
92.26\%
&
$
y(x<0) = \swish\times\elish\times\sinus\text{ and } y(x\geq0)=\relu
$
\\
\midrule
7.
&
92.43\%
&
$
y(x)=\relu
$
\\
8.
&
91.45\%
&
$
y(x) =\elu
$
\\
9.
&
91.43\%
&
$
y(x) = \selu
$
\\
\bottomrule
\end{tabular}
\end{table}

\vspace{-0.5cm}

\begin{table}
\centering
\caption{Performance of top six explored functions for Cifar10-VGG16.}
\label{table:3}
\begin{tabular}{lcl}
\toprule
&
\multicolumn{1}{c}{Accuracy}
&
\multicolumn{1}{c}{Activation Function}
\\
\midrule
1.
&
93.23\%
&
$
y(x)= \elish
$
\\
2.
&
92.89\%
&
$
y(x)= \hardelish 
$
\\
3.
&
92.78\%
&
$
y(x<0) = \swish\times\elish\times\sinus\text{ and } y(x\geq0)=\relu
$
\\
4.
&
91.35\%
&
$
y(x<0) = \sinus\text{ and } y(x\geq0)=\swish+\swish
$
\\
5.
&
91.39\%
&
$
y(x<0) = \hardelish \circ \swish \text{ and } y(x\geq0)~=\swish 
$
\\
6.
&
90.86\%
&
$
y(x) =\swish
$
\\
\midrule
7.
&
93.00\%
&
$
y(x) =\relu
$
\\
8.
&
92.88\%
&
$
y(x) =\selu
$
\\
9.
&
92.60\%
&
$
y(x) =\elu
$
\\
\bottomrule
\end{tabular}
\end{table}

\begin{figure}
\centering
\resizebox{0.9\textwidth}{!}
{
\begin{tabular}{ccc}
    \begin{tikzpicture}[scale=2/3]
\begin{axis}
    [xmin=-4, xmax=3,
    ymin=-1.2, ymax=3,
    x=0.75cm,
    y=0.75cm,
    axis lines=center,
    axis on top=true]

    \addplot [mark=none,draw=red,ultra thick,domain=0:4] {x*1/(1+exp(-x))};
    \addplot [mark=none,draw=red,ultra thick,domain=-4:0] {x*1/(1+exp(-x))};
\end{axis}
\end{tikzpicture} &
    \begin{tikzpicture}[scale=2/3]
\begin{axis}
    [xmin=-6, xmax=3,
    ymin=-1.2, ymax=3,
    x=0.75cm,
    y=0.75cm,
    axis lines=center,
    axis on top=true]

    \addplot [mark=none,draw=red,ultra thick,domain=0:4] {x*1/(1+exp(-x))};
    \addplot [mark=none,draw=red,ultra thick,domain=-10:0] {(max(0, min(1, (x*1/(1+exp(-x)) + 1)/2))*(exp(x*1/(1+exp(-x)))-1)};
\end{axis}
\end{tikzpicture} &
    \begin{tikzpicture}[scale=2/3]
\begin{axis}
    [xmin=-6, xmax=3,
    ymin=-1.2, ymax=3,
    x=0.75cm,
    y=0.75cm,
    axis lines=center,
    axis on top=true]

    \addplot [mark=none,draw=red,ultra thick,domain=0:4] {x};
    \addplot [mark=none,draw=red,ultra thick,domain=-6:0] {(exp(x)-1)*(1/(1+exp(-x)))*sin(deg(x))*(x/(1+exp(-x)))};
\end{axis}
\end{tikzpicture}
    \\ \\
    $y(x) = \swish$&
    $\!\begin{aligned} 
       y(x\geq0) &=\swish \\
       y(x<0) &= \hardelish \circ \swish
    \end{aligned}$ &
    $\!\begin{aligned} 
       y(x\geq0)&=\relu \\
       y(x<0) &= \swish \times \elish \times \sinus
    \end{aligned}$
    \\ \\
\end{tabular}
}

\resizebox{0.9\textwidth}{!}{
\begin{tabular}{ccc}
    \begin{tikzpicture}[scale=2/3]
\begin{axis}
    [xmin=-6, xmax=3,
    ymin=-1.2, ymax=3,
    x=0.75cm,
    y=0.75cm,
    axis lines=center,
    axis on top=true]

    \addplot [mark=none,draw=red,ultra thick,domain=0:4] {x*1/(1+exp(-x))};
    \addplot [mark=none,draw=red,ultra thick,domain=-6:0] {((exp(\x)-1)/(1+exp(-x)))};
\end{axis}
\end{tikzpicture} &
    \begin{tikzpicture}[scale=2/3]
\begin{axis}
    [xmin=-4, xmax=3,
    ymin=-1.2, ymax=3,
    x=0.75cm,
    y=0.75cm,
    axis lines=center,
    axis on top=true]

    \addplot [mark=none,draw=red,ultra thick,domain=0:4] { max(0, min(1, (x+1)/2))*x};
    \addplot [mark=none,draw=red,ultra thick,domain=-10:0] {( exp(x)-1) * max(0, min(1, (x+1)/2))};
\end{axis}
\end{tikzpicture} &
    \begin{tikzpicture}[scale=2/3]
\begin{axis}
    [xmin=-7, xmax=3,
    ymin=-1.2, ymax=3,
    x=0.75cm,
    y=0.75cm,
    axis lines=center,
    axis on top=true]

    \addplot [mark=none,draw=red,ultra thick,domain=0:4] {x*2/(1+exp(-x))};
    \addplot [mark=none,draw=red,ultra thick,domain=-6.28:0] {sin(deg(x))};
    
    
\end{axis}
\end{tikzpicture}\\ \\ 
    $\!\begin{aligned} 
       y(x)&=\elish 
    \end{aligned}$ &
    $\!\begin{aligned} 
       y(x)&=\hardelish 
    \end{aligned}$ &
    $\!\begin{aligned} 
       y(x\geq0) & =\swish+\swish \\
       y(x<0) &= \sinus
    \end{aligned}$
\end{tabular}
}
\caption{Top six explored functions for CIFAR-10.}
\label{plot:cifar10}
\end{figure}
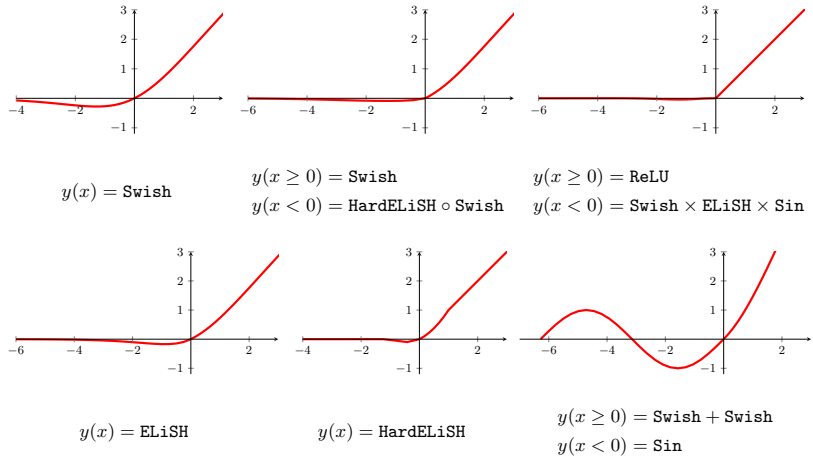

Additionally, we run the same experiment using the VGG-16 framework and show the  
results in \Cref{table:3}. Even though the AFs have not been trained for this 
architecture, we get competitive results: Similarly, we get the best results 
using \elish ($71.25\%$), once again followed by \hardelish. However, as the AFs 
have not been evolved for the VGG architecture, the gap compared to the 
baselines is smaller or even vanishing. For better understanding, we also 
illustrate the top $6$ activation functions for CIFAR-10 in \Cref{plot:cifar10}.


Next, we run the same experiments on CIFAR-100, however, to reduce the 
computational effort, building on a ResNet-20 during evolution. The 
corresponding results for ResNet-56 and VGG-16 are given in \Cref{table:4} and 
\Cref{table:5}. It can be seen in \Cref{table:4} that \elish ($74.65\%$) and the 
compositional function of {$\max(\sinus, \hardelish)$} and \sinus  in the 
negative part and \swish in the positive part ($74.31\%$) show the best 
performances for ResNet56. For VGG, as demonstrated in \Cref{table:5}, the 
activation function consisting of \hardelish in the negative part and (\selu+ 
\linear) in the positive part ($71.25\%$) and \swish ($71.23\%$) yield the best 
results. These results show that for negative inputs \hardelish, \sinus and the 
combinations of them come up during evolution. In
\Cref{plot:cifar100}, also for CIFAR-100 we show an illustration of the top $6$ 
evolved activation functions.

\begin{table}
\centering
\caption{Performance of top six explored functions for CIFAR100-ResNet56.}
\label{table:4}
\begin{tabularx}{\textwidth}{lcm{12cm}}
\toprule
&
\multicolumn{1}{c}{Accuracy}
&
\multicolumn{1}{c}{Activation Function}
\\
\midrule
1.
&
74.65\%
&
$
y(x)= \elish
$
\\
2.
&
74.31\%
&
$
y(x<0)=\max(\sinus,\hardelish)\circ \sinus \text{ and } y(x\geq0)~=\swish 
$
\\

3.
&
74.09\%
&
$
y(x<0) = \sinus\text{ and } y(x\geq0)=\max(\selu,\selu+\linear)\circ \relu
$
\\
4.
&
74.05\%
&
$
y(x<0) = \hardelish\text{ and } y(x\geq0)=\max(\selu,\selu+\linear)
$
\\
5.
&
73.98\%
&
$
y(x)= \swish
$
\\
6.
&
73.61\%
&
$
y(x<0) = \max(\sinus,\hardelish ) \text{ and } y(x\geq0)~= \max(\selu,\selu+\linear)
$
\\

\midrule
7.
&
73.31\%
&
$
y(x)=\relu
$
\\
8.
&
72.58\%
&
$
y(x) =\elu
$
\\
9.
&
71.57\%
&
$
y(x) = \selu
$
\\
\bottomrule
\end{tabularx}
\end{table}

\begin{table}
\centering
\caption{Performance of top six explored functions for CIFAR100-VGG16.}
\label{table:5}
\begin{tabularx}{\textwidth}{lcm{12cm}}
\toprule
&
\multicolumn{1}{c}{Accuracy}
&
\multicolumn{1}{c}{Activation Function}
\\
\midrule
1.
&
71.25\%
&
$
y(x<0) = \hardelish\text{ and } y(x\geq0)=\max(\selu,\selu+\linear)
$
\\
2.
&
71.23\%
&
$
y(x)= \swish
$
\\
3.
&
70.80\%
&
$
y(x<0) = \max(\sinus,\hardelish ) \text{ and } y(x\geq0)~= \max(\selu,\selu+\linear)
$
\\

4.
&
70.77\%
&
$
y(x<0)=\max(\sinus,\hardelish)\circ \sinus \text{ and } y(x\geq0)~=\swish 
$
\\
5.
&
70.74\%
&
$
y(x<0) = \sinus\text{ and } y(x\geq0)=\max(\selu,\selu+\linear)\circ \relu
$
\\
6.
&
70.70\%
&
$
y(x) = \elish
$
\\

\midrule
7.
&
71.12\%
&
$
y(x)=\elu
$
\\
8.
&
70.74\%
&
$
y(x) =\relu
$
\\
9.
&
70.59\%
&
$
y(x) = \selu
$
\\
\bottomrule
\end{tabularx}
\end{table}


Finally, we run the same experiments on Tiny ImageNet using the same setup as 
used for CIFAR-100. The thus obtained results for ResNet-56 and VGG-16 are given 
in \Cref{table:6} and \Cref{table:7}. It can be seen from \Cref{table:6} that 
the activation function with the combination \hardelish in the negative part and  
{$\min(\elu,\swish)$} in the positive part ($57.53\%$) and \elish ($57.34\%$) 
demonstrate the best performances. As can be seen from \Cref{table:7}, \elish 
also provides good results for VGG16 ($52.30\%$). It seems that that \swish and 
\linear (and a combination of them) in the positive part are the best fitting 
activation functions for this dataset. In addition, we show an illustration of 
the top $6$ explored activation functions in \Cref{plot:tinyImageNet}.

\begin{table}[h]
\centering
\caption{Performance of top six explored functions for ImageNet-Resnet56.}
\label{table:6}
\begin{tabularx}{\textwidth}{lcm{10cm}}
\toprule
&
\multicolumn{1}{c}{Accuracy}
&
\multicolumn{1}{c}{Activation Function}
\\
\midrule

1.
&
57.53\%
&
$
y(x<0) = \hardelish\text{ and } y(x\geq0)=\min(\elu,\swish)
$
\\
2.
&
57.34\%
&
$
y(x) =\elish
$
\\
3.
&
57.07\%
&
$
y(x) =\swish
$
\\

4.
&
56.68\%
&
$
y(x<0) = \min(\elish,\relu) \text{ and } y(x\geq0)=(\relu +\elish)
$
\\
5.
&
56.62\%
&
$
y(x<0) = \hardelish \text{ and } y(x\geq0)=\elu
$
\\
6.
&
56.32\%
&
$
y(x<0)= \swish +\min(\elish:\relu) \text{ and } y(x\geq0)=( \relu+\elish)
$
\\


\midrule
7.
&
57.27\%
&
$
y(x)=\relu
$
\\
8.
&
55.32\%
&
$
y(x) =\elu
$
\\
9.
&
50.09\%
&
$
y(x) = \selu
$
\\
\bottomrule
\end{tabularx}
\end{table}

\vspace{-0.5cm}

\begin{table}[h]
\centering
\caption{Performance of top six explored functions for ImageNet-VGG.}
\label{table:7}
\begin{tabularx}{\textwidth}{lcm{10cm}}
\toprule
&
\multicolumn{1}{c}{Accuracy}
&
\multicolumn{1}{c}{Activation Function}
\\
\midrule

1.
&
52.30\%
&
$
y(x) =\elish
$
\\
2.
&
52.19\%
&
$
y(x) =\swish
$
\\

3.
&
51.77\%
&
$
y(x<0) = \min(\elish,\relu) \text{ and } y(x\geq0)=(\relu +\elish)
$
\\

4.
&
51.48\%
&
$
y(x<0) = \hardelish \text{ and } y(x\geq0)~= \min(\elu,\swish)
$
\\
5.
&
51.41\%
&
$
y(x<0)= \swish +\min(\elish,\relu) \text{ and } y(x\geq0)=(\relu+\elish)
$
\\
6.
&
51.34\%
&
$
y(x<0) = \hardelish \text{ and } y(x\geq0)=\elu
$
\\
\midrule
7.
&
51.70\%
&
$
y(x)=\relu
$
\\
8.
&
51.10\%
&
$
y(x) =\elu
$
\\
9.
&
49.54\%
&
$
y(x) = \selu
$
\\
\bottomrule
\end{tabularx}
\end{table}

\subsection{Discussion}

The results presented above clearly show that for three different problems, 
totally different activation functions are estimated. In fact, from the obtained 
results we can recognize two different kinds of activation functions considering 
their positive part only: (a) non expansion (contraction) and (b) expansion 
mapping, which can be considered as a function approximator \cite{mapping}. In 
general, an expansion mapping expands changes in input values, while a 
contraction mapping is less sensitive to changes in input values.
  
For CIFAR-10, with exception of activation function $\#3$ (\Cref{table:2}) 
all evolved activation functions for positive inputs are contraction mappings. 
This can be explained by the fact that some classes build multi-modal clusters, 
resulting in large intra-class distances, which might cause misclassifications. 
A contraction mapping like \swish, however, helps to reduce this effect, 
improving the classification accuracy.
   
For CIFAR-100, in contrast, we observe more expanding mappings. If the instances 
of a class are close to each other, this allows for better exploiting  
discriminative power. The expanding character of activation functions can 
especially be recognized from repetitive piecewise functions in the negative 
part, as can be seen from \Cref{plot:cifar100}.  
   
For Tiny ImageNet, a more complex benchmark, expansion and non expansion 
mappings can be recognized. This can, again, be seen from the negative part of 
the activation functions. When the rate decay on negative side is exponential 
like in activation function $\#4$ (\Cref{table:6}) 
the tendency is toward expanding, otherwise it is identity.


\vspace{1.5cm}


\clearpage


\begin{figure}[!h]
\centering
\resizebox{0.9\textwidth}{!}
{
\begin{tabular}{cccc}
     \begin{tikzpicture}[scale=2/3]
\begin{axis}
    [xmin=-4, xmax=3,
    ymin=-1.2, ymax=3,
    x=0.75cm,
    y=0.75cm,
    axis lines=center,
    axis on top=true]
    \addplot [mark=none,draw=red,ultra thick,domain=0:4] {x*1/(1+exp(-x))};
    \addplot [mark=none,draw=red,ultra thick,domain=-10:0] {(exp(x)-1)*1/(1+exp(-x))};
\end{axis}
\end{tikzpicture} &
    \begin{tikzpicture}[scale=2/3]
\begin{axis}
    [xmin=-4, xmax=3,
    ymin=-1.2, ymax=3,
    x=0.75cm,
    y=0.75cm,
    axis lines=center,
    axis on top=true]

    \addplot [mark=none,draw=red,ultra thick,domain=0:4] {x*1/(1+exp(-x))};
    \addplot [mark=none,draw=red,ultra thick,domain=-4:0] {x*1/(1+exp(-x))};
\end{axis}
\end{tikzpicture} &
    \begin{tikzpicture}[scale=2/3]
\begin{axis}
    [xmin=-4, xmax=3,
    ymin=-1.2, ymax=3,
    x=0.75cm,
    y=0.75cm,
    axis lines=center,
    axis on top=true]

    \addplot [mark=none,draw=red,ultra thick,domain=0:4] {max(x,sin(deg(x)))};
    \addplot [mark=none,draw=red,ultra thick,domain=-10:0] {(exp(x)-1)*(1/(1+exp(-x)))};
\end{axis}
\end{tikzpicture} &
    \begin{tikzpicture}[scale=2/3]
\begin{axis}
    [xmin=-7, xmax=3,
    ymin=-1.2, ymax=3,
    x=0.75cm,
    y=0.75cm,
    axis lines=center,
    axis on top=true]

    \addplot [mark=none,draw=red,ultra thick,domain=0:4] {max(1.0507009873554804934193349852946*x,x+1.0507009873554804934193349852946*x)};
    \addplot [mark=none,draw=red,ultra thick,domain=-6.28:0] {sin(deg(x))};
    
    
\end{axis}
\end{tikzpicture} 
    \\ \\
     $y(x)=\elish$ &
    $y(x)=\swish$ &
    $\!\begin{aligned} 
       y(x\geq0)&=\max(\linear,\sinus) \\
       y(x<0) &= \elish
    \end{aligned}$ &
    $\!\begin{aligned} 
       y(x\geq0)&=\max(\selu,\selu+\linear) \\
       y(x<0) &= \sinus
    \end{aligned}$ 
    \\ \\
\end{tabular}
}
\resizebox{0.9\textwidth}{!}{
\begin{tabular}{ccc}
    \begin{tikzpicture}[scale=2/3]
\begin{axis}
    [xmin=-12.56, xmax=3,
    ymin=-1.2, ymax=3,
    x=0.5cm,
    y=0.5cm,
    axis lines=center,
    axis on top=true]

    \addplot [mark=none,draw=red,ultra thick,domain=0:4] {max(1.0507009873554804934193349852946*x,x+1.0507009873554804934193349852946*x)};
    \addplot [mark=none,draw=red,ultra thick,domain=-12.56:0] {
    max(
        sin(deg(x)),
        ( exp(x)-1) * max(0, min(1, (x+1)/2)) )
        )
    };
\end{axis}
\end{tikzpicture} &
    \begin{tikzpicture}[scale=2/3]
\begin{axis}
    [xmin=-12.56, xmax=3,
    ymin=-1.2, ymax=3,
    x=0.5cm,
    y=0.5cm,
    axis lines=center,
    axis on top=true]

    \addplot [mark=none,draw=red,ultra thick,domain=0:4] {x*1/(1+exp(-x))};
    \addplot [mark=none,draw=red,ultra thick,domain=-12.56:0] {
    max(
        sin(deg(sin(deg(x)))),
        ( exp(sin(deg(x)))-1) * max(0, min(1, (sin(deg(x))+1)/2)) )
        )
    };
\end{axis}
\end{tikzpicture} \\
    $\!\begin{aligned} 
       y(x\geq0) &=\max(\selu, \selu+\linear) \\
       y(x<0) &= \max(\sinus, \hardelish)
    \end{aligned}$&
    $\!\begin{aligned} 
       y(x\geq0) & = \swish\\
       y(x<0) &= \max(\sinus,\hardelish) \circ \sinus
    \end{aligned}$
\end{tabular}
}
\caption{Top six explored functions for CIFAR-100.}
\label{plot:cifar100}
\end{figure}

\vspace{-0.75cm}


\begin{figure}[!h]
\centering
\resizebox{0.9\textwidth}{!}
{
\begin{tabular}{ccc}
    \begin{tikzpicture}[scale=2/3]
\begin{axis}
    [xmin=-4, xmax=3,
    ymin=-1.2, ymax=3,
    x=0.75cm,
    y=0.75cm,
    axis lines=center,
    axis on top=true]
    \addplot [mark=none,draw=red,ultra thick,domain=0:4] {min(\x,x*1/(1+exp(-x)))};
    \addplot [mark=none,draw=red,ultra thick,domain=-10:0] {( exp(x)-1) * max(0, min(1, (x+1)/2))};
\end{axis}
\end{tikzpicture} &
    \begin{tikzpicture}[scale=2/3]
\begin{axis}
    [xmin=-4, xmax=3,
    ymin=-1.2, ymax=3,
    x=0.75cm,
    y=0.75cm,
    axis lines=center,
    axis on top=true]
    \addplot [mark=none,draw=red,ultra thick,domain=0:4] {\x+(x*1/(1+exp(-x)))};
    \addplot [mark=none,draw=red,ultra thick,domain=-10:0] {min(0,(exp(x)-1)*1/(1+exp(-x)))};
\end{axis}
\end{tikzpicture} &
    \begin{tikzpicture}[scale=2/3]
\begin{axis}
    [xmin=-4, xmax=3,
    ymin=-1.2, ymax=3,
    x=0.75cm,
    y=0.75cm,
    axis lines=center,
    axis on top=true]
    \addplot [mark=none,draw=red,ultra thick,domain=0:4] {\x};
    \addplot [mark=none,draw=red,ultra thick,domain=-10:0] {( exp(x)-1) * max(0, min(1, (x+1)/2))};
\end{axis}
\end{tikzpicture} 
    \\ \\
    $\!\begin{aligned} 
       y(x\geq0)&=\min(\elu,\swish) \\
       y(x<0) &= \hardelish
    \end{aligned}$ &
    $\!\begin{aligned} 
       y(x\geq0)&=\relu+\elish \\
       y(x<0) &= \min(\elish,\relu)
    \end{aligned}$ &
    $\!\begin{aligned} 
       y(x\geq0)&=\elu \\
       y(x<0) &= \hardelish
    \end{aligned}$
\end{tabular}
}

\resizebox{0.9\textwidth}{!}{
\begin{tabular}{ccc}
    \begin{tikzpicture}[scale=2/3]
\begin{axis}
    [xmin=-4, xmax=3,
    ymin=-1.2, ymax=3,
    x=0.75cm,
    y=0.75cm,
    axis lines=center,
    axis on top=true]
    \addplot [mark=none,draw=red,ultra thick,domain=0:4] {x*1/(1+exp(-x))};
    \addplot [mark=none,draw=red,ultra thick,domain=-10:0] {(exp(x)-1)*1/(1+exp(-x))};
\end{axis}
\end{tikzpicture} &
    \begin{tikzpicture}[scale=2/3]
\begin{axis}
    [xmin=-4, xmax=3,
    ymin=-1.2, ymax=3,
    x=0.75cm,
    y=0.75cm,
    axis lines=center,
    axis on top=true]
    \addplot [mark=none,draw=red,ultra thick,domain=0:4] {\x+(x*1/(1+exp(-x)))};
    \addplot [mark=none,draw=red,ultra thick,domain=-10:0] {x*1/(1+exp(-x))+min(0,(exp(x)-1)*1/(1+exp(-x)))};
\end{axis}
\end{tikzpicture} &
   \begin{tikzpicture}[scale=2/3]
\begin{axis}
    [xmin=-4, xmax=3,
    ymin=-1.2, ymax=3,
    x=0.75cm,
    y=0.75cm,
    axis lines=center,
    axis on top=true]

    \addplot [mark=none,draw=red,ultra thick,domain=0:4] {x*1/(1+exp(-x))};
    \addplot [mark=none,draw=red,ultra thick,domain=-4:0] {x*1/(1+exp(-x))};
\end{axis}
\end{tikzpicture} \\
    $y(x)=\elish$ &
    $\!\begin{aligned} 
       y(x\geq0) &=\relu+\elish \\
       y(x<0) &= \swish + \min(\elish,\relu) \\
    \end{aligned}$&
    $\!\begin{aligned} 
       y(x) & = \swish
       
    \end{aligned}$
\end{tabular}
}
\caption{Top six explored functions for TinyImageNet.}
\label{plot:tinyImageNet}
\end{figure}
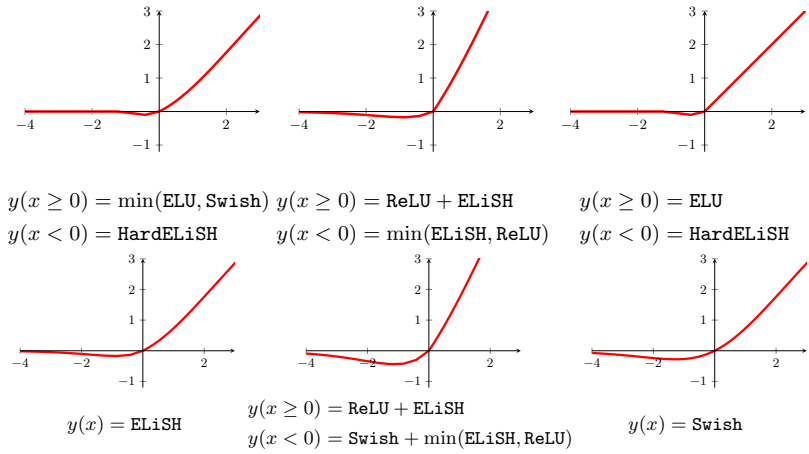

\vspace{-1.0cm}
\section{Conclusion and Future Work}
\label{sec:conclusion}

Even though deep learning approaches allow  end-to-end learning for a variety of 
applications, there are still many parameters which need do be manually set. An 
important parameter, which is often ignored, is the choice of activation 
functions. Thus, we tackled this problem and studied the importance of 
activation functions when learning DNN models for classification. In particular, 
our contribution is threefold. First, we introduced two new activation functions 
based on theoretical considerations. Second, we introduced a genetic algorithm 
based evolving procedure to learn the best activation function for a given task. 
Third, we gave a detailed evaluation and a discussion on our findings. Future 
work will include a more thorough experimental evaluation and an analysis of the 
effect of activation functions on the computational complexity during DNN 
learning.

\bibliographystyle{splncs}
\bibliography{_abbrv_,ref}

\end{document}